\documentclass[conference]{IEEEtran}
\usepackage[T1]{fontenc}
\usepackage[utf8]{inputenc}
\usepackage{cite}
\usepackage{url}
\usepackage{array}
\usepackage{booktabs}
\usepackage{tabularx}
\usepackage{amsmath}
\usepackage{enumitem}
\usepackage{xcolor}
\usepackage[hidelinks]{hyperref}

\title{From Governance Norms to Enforceable Controls: A Layered Translation Method for Runtime Guardrails in Agentic AI}

\author{%
\IEEEauthorblockN{Christopher Koch}
\IEEEauthorblockA{Independent Researcher}
}

\begin{document}
\maketitle

\begin{abstract}
Agentic AI systems plan, use tools, maintain state, and produce multi-step trajectories with external effects. Those properties create a governance problem that differs materially from single-turn generative AI: important risks emerge during execution, not only at model development or deployment time. Governance standards such as ISO/IEC 42001, ISO/IEC 23894, ISO/IEC 42005, ISO/IEC 5338, ISO/IEC 38507, and the NIST AI Risk Management Framework are therefore highly relevant to agentic AI, but they do not by themselves yield implementable runtime guardrails. This paper proposes a layered translation method that connects standards-derived governance objectives to four control layers: governance objectives, design-time constraints, runtime mediation, and assurance feedback. It distinguishes governance objectives, technical controls, runtime guardrails, and assurance evidence; introduces a control tuple and runtime-enforceability rubric for layer assignment; and demonstrates the method in a procurement-agent case study. The central claim is modest: standards should guide control placement across architecture, runtime policy, human escalation, and audit, while runtime guardrails are reserved for controls that are observable, determinate, and time-sensitive enough to justify execution-time intervention.
\end{abstract}


\section{Introduction}
Large language model (LLM) systems are increasingly embedded in agentic applications that can decompose tasks, invoke tools, preserve memory, coordinate with external services, and generate long action sequences with limited human intervention. This transition changes the control problem. A conventional generative model can often be assessed at the level of prompts, outputs, and offline evaluation. By contrast, an agent may look harmless at each individual step while still producing an unacceptable trajectory when its actions are composed over time \cite{pathpolicies,mi9}.

Organizations already have a substantial governance baseline. ISO/IEC 42001 provides an AI management system framework; ISO/IEC 23894 addresses AI risk management; ISO/IEC 42005 structures AI impact assessment; ISO/IEC 5338 covers lifecycle processes; ISO/IEC 38507 addresses governance implications for organizations; and NIST provides the AI RMF, the Generative AI Profile, and a dedicated AI Agent Standards Initiative \cite{iso42001,iso23894,iso42005,iso5338,iso38507,nistairmf,nistgaiprofile,nistagentinitiative}. These instruments provide governance intent, risk structure, and accountability baselines for agentic AI, but not executable runtime policy. The real question is therefore not whether standards can be compiled directly into guardrails, but how standards-derived objectives should be translated across design-time, runtime, and assurance layers \cite{policyasprompt,policycards}.

This paper argues for that narrower and more useful claim. It makes three contributions:
\begin{enumerate}[leftmargin=*, itemsep=0.15em]
    \item It distinguishes governance objectives, technical controls, runtime guardrails, and assurance evidence as different artifacts with different roles.
    \item It proposes a governance-to-control translation method centered on an explicit control tuple, a runtime-enforceability rubric, and layer assignment.
    \item It demonstrates the method with a procurement-agent case study and derives an evaluation agenda grounded in recent runtime-governance and agent-safety literature.
\end{enumerate}

\section{Scope and Method}
This paper offers a design-oriented interpretive framework rather than an empirical benchmark or a clause-by-clause compliance mapping. It draws on public ISO and NIST descriptions of the relevant governance frameworks \cite{iso42001,iso23894,iso42005,iso5338,iso38507,nistairmf,nistgaiprofile,nistagentinitiative,nistidentity} and on recent literature on runtime governance, agent guardrails, and agent evaluation \cite{mi9,policyasprompt,policycards,pathpolicies,guardrailsurvey,foundationalguardrail,agentsafetybench,webguard,gap,toolsafe,agentdog,proofguardrail,abcbench}. The question is simple: \emph{what control architecture follows when standards-derived governance objectives are applied to agentic systems?} Here, \emph{runtime guardrail} refers narrowly to execution-time mechanisms that allow, deny, delay, escalate, or reshape actions based on policy, context, identity, or trajectory rather than to governance documents, organizational processes, or post hoc audit alone. Negative claims about what standards do not specify are therefore claims about public descriptions and governance logic, not formal proofs of absence, and the recent guardrail literature is used as directional evidence rather than settled validation. The placement logic is also informed by classic work on enforceable security policies \cite{schneider}.

\section{Related Work and Gap}
Relevant work now exists in four streams.

\textbf{Standards and governance frameworks.} ISO and NIST now provide serious AI governance baselines, but these documents remain technology-agnostic by design and emphasize management systems, risk processes, impact assessment, lifecycle discipline, and organizational accountability rather than executable control logic \cite{iso42001,iso23894,iso42005,iso5338,iso38507,nistairmf,nistgaiprofile,nistagentinitiative,nistidentity}.

Comparable patterns appear in high-assurance software and autonomy engineering. NASA emphasizes classification, tailoring, traceability, IV\&V, and objective evidence, while NIST SP 800-160 Vol. 2 and DARPA's Assured Autonomy frame trust and resilience as life-cycle and continual-assurance problems rather than collections of isolated runtime checks \cite{nasa7150,nasa87398,nasahdbk2203,nist800160v2,darpaassured}.

\textbf{Runtime governance and policy compilation.} Recent papers increasingly argue that agentic systems need runtime oversight. MI9 frames agent governance as an integrated runtime problem \cite{mi9}. Policy-as-Prompt explores turning policy and design artifacts into runtime guardrails \cite{policyasprompt}. Policy Cards proposes a machine-readable deployment-layer representation of operational constraints \cite{policycards}. Policies on Paths goes further by formalizing compliance as a function of partial execution paths rather than isolated prompts \cite{pathpolicies}.

\textbf{Agent safety and action-level guardrails.} Empirical work also shows that action safety is a distinct problem. Agent-SafetyBench reports that none of the tested agents exceeds a safety score of 60\% \cite{agentsafetybench}. WebGuard finds frontier models below 60\% accuracy in predicting action outcomes and below 60\% recall on high-risk web actions without specialized safeguards \cite{webguard}. Mind the GAP shows that text-level safety does not reliably transfer to tool-call safety \cite{gap}. ToolSafe demonstrates that step-level guardrails can reduce harmful tool invocations by 65\% on average under attack while improving benign task completion by roughly 10\% \cite{toolsafe}. AgentDoG and Proof-of-Guardrail extend the discussion to richer diagnostics and verifiable execution claims \cite{agentdog,proofguardrail}. Foundational Guardrail argues that pre-execution intervention can be safer than purely post-execution filtering for general agentic systems \cite{foundationalguardrail}.

\textbf{Evaluation rigor.} Benchmark methodology itself remains fragile. Best-practice work on agentic benchmarks shows that flawed reward design and task setup can materially distort measured performance \cite{abcbench}.

Taken together, this literature still leaves an important practical gap: organizations need a disciplined way to decide which standards-derived requirements should be enforced at runtime, which should remain design-time constraints, which require human escalation, and which are best handled as assurance obligations. This paper addresses that gap with a compact translation method rather than a new benchmark or a new standard.

\section{Why Direct Translation Is Insufficient}
\subsection{Category Mismatch}
ISO/IEC 42001, ISO/IEC 23894, ISO/IEC 42005, ISO/IEC 5338, ISO/IEC 38507, and the NIST AI RMF are frameworks for management, risk, lifecycle, impact, and governance \cite{iso42001,iso23894,iso42005,iso5338,iso38507,nistairmf}. By themselves they do not define a policy language, an action schema, or an execution model. Treating them as direct guardrail specifications conflates governance objectives with technical mechanisms, and an organization can satisfy a governance framework while its deployed agent still lacks meaningful runtime controls.

\subsection{Limited Runtime Enforceability}
Some governance norms translate relatively well into technical controls: least privilege, scoped authorization, logging, approval gates, or retention limits. Others do not. Requirements related to fairness, proportionality, human acceptability, or societal impact require contextual judgment that cannot be safely reduced to a deterministic runtime rule without substantial normative simplification, and recent policy-compilation work still depends on interpretation, provenance, and human oversight rather than naive automation \cite{policyasprompt,policycards}.

\subsection{Guardrails Are Fallible}
Dong et al.'s survey of LLM safeguards shows that even non-agentic safety mechanisms are layered, context-sensitive, and incomplete \cite{guardrailsurvey}. For agents, the problem is harder because decisions unfold across tool calls and trajectories rather than one-shot outputs. AgentDoG, WebGuard, ToolSafe, and Agent-SafetyBench collectively show both the need for action-level controls and the gap between current guardrail performance and high-stakes reliability \cite{agentdog,agentsafetybench,webguard,toolsafe}, while Proof-of-Guardrail shows that even proving guardrail execution is not the same as proving safety \cite{proofguardrail}.

\subsection{Control-Layer Misplacement}
A poorly scoped agent with broad tool access cannot usually be made safe by runtime filtering alone. Huang et al.\ argue that pre-execution intervention is often safer than post-execution filtering because some harms become irreversible once actions execute \cite{foundationalguardrail}. More broadly, some controls belong in architecture, model choice, network isolation, human workflow design, and post-deployment assurance rather than live policy checks alone; broader assured-autonomy work makes the same point by treating runtime assurance as one element within a larger design-time and operation-time assurance regime \cite{nistgaiprofile,nistidentity,nist800160v2,darpaassured}.

The implication is not that runtime guardrails are unimportant. It is that they should be treated as one layer in a broader governance system rather than as the entire operational meaning of governance.

\section{Governance-to-Control Translation Method}
The proposed method has five steps. It is intentionally lightweight so that it can be used as a design-review tool rather than only as a research abstraction.

\subsection{Different Artifacts Need Different Layers}
A stronger argument begins by separating four kinds of artifacts that are often conflated.

\textbf{Governance objective:} a normative goal such as accountability, least privilege, impact awareness, or risk reduction, often sourced from a standard, regulation, or internal policy.

\textbf{Technical control:} a mechanism intended to operationalize some aspect of that objective, such as scoped credentials, approval gates, logging, or anomaly detection.

\textbf{Runtime guardrail:} a subset of technical controls that intervene during execution by allowing, denying, delaying, escalating, or reshaping actions.

\textbf{Assurance evidence:} artifacts used to demonstrate what controls exist, whether they executed, and with what effect, such as logs, signed attestations, audit traces, incident reports, or validation records.

\subsection{Step 1: Extract the Normative Objective}
Begin with a standards-derived statement or organizational policy objective. Example forms include ``ensure access is authorized and auditable,'' ``assess impacts on affected parties,'' or ``maintain continual monitoring and improvement'' \cite{iso42001,iso42005,nistairmf}.

\subsection{Step 2: Normalize It into a Control Tuple}
Any candidate control should be rewritten into a structured tuple
\begin{equation}
\kappa = \langle a, x, r, \phi, \delta, \epsilon, o \rangle,
\end{equation}
where $a$ is the acting principal (human, agent, or sub-agent), $x$ is the action class, $r$ is the protected resource or external effect, $\phi$ is the precondition or relevant context, $\delta$ is the control decision (allow, deny, escalate, log-only, or rewrite), $\epsilon$ is the evidence artifact to be produced, and $o$ is the accountable owner. This makes the proposed control concrete enough to inspect, compare, and audit. For example, ``only approved vendors may receive purchase orders'' can be normalized as a procurement-agent action to create a purchase order against a vendor record under the precondition that the vendor is on an approved list, with a runtime allow-or-deny decision, a signed decision trace, and procurement ownership.

\subsection{Step 3: Score Runtime Enforceability}
Next, assess whether the objective is actually suitable for runtime enforcement. A control should be considered a strong runtime candidate only when the protected event is observable before execution, the decision rule is sufficiently determinate, the intervention is operationally tolerable, and post hoc review would be too late. Table~\ref{tab:rubric} summarizes the rubric.

\begin{table}[t]
\caption{Runtime-enforceability rubric}
\label{tab:rubric}
\centering
\footnotesize
\begin{tabularx}{\columnwidth}{@{}>{\raggedright\arraybackslash}p{0.19\columnwidth}>{\raggedright\arraybackslash}X>{\raggedright\arraybackslash}X@{}}
\toprule
Criterion & High runtime-enforceability & Low runtime-enforceability \\
\midrule
Timing of harm & Harm must be prevented before execution & Harm is mainly evaluable after the fact \\
Pre-action observability & Required state and context are machine-observable & Critical context is absent or only discoverable later \\
Rule determinacy & Policy can be written as a crisp operational rule & Policy requires open-ended interpretation or balancing \\
Judgment load & Limited social or ethical judgment required & Human or contextual judgment is central \\
Reversibility & Mistakes are hard to undo; intervention is urgent & Action can be audited or corrected later \\
Evidence clarity & Control outcomes can be logged and attributed cleanly & Evidence is ambiguous or weakly attributable \\
\bottomrule
\end{tabularx}
\end{table}

These criteria are heuristic rather than exhaustive, but they follow from the failure modes above and from classic work on enforceable security policies: a control cannot reliably operate online when the trigger is unobservable, the protected object is too diffuse, the required context is unavailable, or the intervention itself is operationally untenable \cite{schneider}. For agentic systems, this runtime layer usually attaches to the orchestrator or tool-dispatch boundary, where actions can still be inspected before external effects occur.

\subsection{Step 4: Assign the Primary Control Layer}
The objective is then assigned to one or more layers:
\begin{enumerate}[leftmargin=*, itemsep=0.1em]
    \item \textbf{Governance objective layer:} normative intent, ownership, thresholds, and exceptions.
    \item \textbf{Design-time layer:} architecture, least-privilege scoping, tool exposure, dataset and prompt boundaries, sandbox design.
    \item \textbf{Runtime layer:} action validation, policy checks, approval gates, dynamic authorization, anomaly detection, and containment.
    \item \textbf{Assurance layer:} telemetry, audits, incident review, attestation, and performance or drift monitoring.
\end{enumerate}
Human escalation cuts across the stack and is the default destination for ambiguous, high-impact, or low-determinacy decisions.

\subsection{Step 5: Specify Evidence and Ownership}
Every translated control must name both the evidence artifact and the accountable owner. A runtime rule without attributable logs or ownership is hard to audit; an assurance claim without evidence is merely a promise \cite{proofguardrail}.

\section{Worked Case Study: Enterprise Procurement Agent}
To make the method concrete, consider an enterprise procurement agent that can search approved vendor catalogs, retrieve contract data, compare quotes, draft purchase orders, and send requests for approval or vendor communication. The example is simple but realistic enough to stress path dependence, authorization, and auditability.

The organization defines five governance requirements inspired by standards and internal policy. Table~\ref{tab:case} shows how they translate across layers. The monetary threshold is illustrative and organization-specific; it is not implied by ISO or NIST.

\begin{table*}[t]
\caption{Worked case study: mapping procurement-agent requirements to layers}
\label{tab:case}
\centering
\footnotesize
\begin{tabularx}{\textwidth}{@{}>{\raggedright\arraybackslash}p{0.23\textwidth}>{\raggedright\arraybackslash}p{0.13\textwidth}>{\raggedright\arraybackslash}X>{\raggedright\arraybackslash}p{0.19\textwidth}@{}}
\toprule
Requirement & Runtime enforceability & Primary implementation layer(s) & Example evidence artifact \\
\midrule
Only approved vendors may receive purchase orders & High & Design-time vendor-directory scoping + runtime vendor-ID allowlist check before PO creation & Signed action trace showing vendor lookup, policy decision, and PO event \\
Purchases above EUR~5{,}000 require human approval & High & Runtime approval gate with delegated identity and threshold check & Approval token, requester identity, timestamp, and immutable approval log \\
The agent may access only systems necessary for procurement tasks & High to medium & Design-time least-privilege credential scoping + runtime authorization for specific tools and actions & Issued scopes, access logs, denied-action logs \\
Supplier ranking should remain fair, explainable, and contestable & Low & Design-time ranking design + assurance audit + human review for exceptions & Periodic audit report, explanation template, exception register \\
All state-changing actions must be attributable and replayable & Medium to high & Runtime telemetry + assurance retention and replay pipeline & End-to-end trace with actor, tool calls, arguments, results, and policy outcomes \\
\bottomrule
\end{tabularx}
\end{table*}

This example clarifies the central thesis. The first three requirements are good runtime candidates because the relevant state is available before execution and the rules are crisp. The fourth is not a strong runtime candidate because ``fair and contestable'' is too open-ended to encode safely as a deterministic pre-action check. The fifth spans runtime and assurance because logging must happen during execution, but replayability and review are post hoc functions.

The case study also reveals an architectural point: runtime guardrails work best when design-time scoping has already constrained the action space. A vendor allowlist is far easier to enforce if the agent is exposed only to procurement tools and limited credentials in the first place, consistent with classic least-privilege design principles \cite{nistidentity,policycards,saltzer}.

\section{Evaluation Criteria}
The method should be evaluated along five dimensions drawn directly from the literature.

\subsection{Policy Fidelity}
Does the translated control preserve the meaning of the original governance objective, or has it narrowed a broad requirement into a misleading heuristic? This is especially important for standards-derived controls that mix legal, organizational, and technical language \cite{policyasprompt,policycards}.

\subsection{Intervention Quality}
For runtime controls, what are the precision and recall on harmful actions, and what benign actions are wrongly blocked? WebGuard and ToolSafe show both the need for action-level intervention and the difficulty of achieving adequate accuracy \cite{webguard,toolsafe}.

\subsection{Trajectory Coverage}
Can the control reason over partial paths, delegated sub-agents, tool arguments, and accumulated state, or does it only moderate isolated prompts? Path dependence is central for agentic governance \cite{pathpolicies,agentdog}.

\subsection{Safety--Utility Trade-off}
What latency, task-completion loss, false-escalation burden, or additional human review load is introduced by the controls? Runtime safety that destroys usability will be bypassed in practice \cite{guardrailsurvey}.

\subsection{Evidence Completeness}
Can a third party later determine which policy fired, whether the declared guardrail actually executed, and which human or service owned the decision? Proof-of-Guardrail shows that even execution claims may require verification mechanisms \cite{proofguardrail}.

A practical evaluation program should combine these dimensions with benchmark hygiene. Weak benchmark design can materially distort perceived safety or capability improvements \cite{abcbench}, and Mind the GAP shows that text-level refusal behavior is not an adequate proxy for tool-call safety \cite{gap}. For the layered model proposed here, evaluation should also ask whether controls were assigned to the right layer in the first place: a badly placed runtime rule is a design failure even if it executes correctly.

\section{Limitations and Threats to Validity}
This paper is still a position-plus-method paper, not an empirical standards implementation. It relies on public scope statements and summaries for the ISO documents rather than exhaustive clause-by-clause interpretation \cite{iso42001,iso23894,iso42005,iso5338,iso38507}; much of the agent-guardrail literature is recent and includes preprints \cite{mi9,policyasprompt,pathpolicies,webguard,toolsafe}; and the method does not claim that compliance can be inferred solely from runtime traces. Sector-specific law, organizational process, and human judgment remain indispensable. These limitations narrow the claim: the contribution is a disciplined design method for translating governance intent into layered controls, not a complete compliance framework.

\section{Conclusion}
Directly compiling ISO and NIST standards into runtime guardrails is too strong. Standards define governance intent, management expectations, and risk questions; runtime guardrails are only one family of mechanisms for operationalizing those goals. The practical implication is simple: each control should be placed in the layer best suited to enforce it. Runtime guardrails matter most where events are observable, rules are crisp, and intervention must occur before harm; elsewhere, architecture, review, and assurance should carry the load.

This paper contributes a concrete translation method, a runtime-enforceability rubric, and a worked case study showing how standards-informed requirements can be assigned to design-time constraints, runtime guardrails, human escalation, and assurance evidence.

Future work should validate the method empirically on domain-specific agent deployments and measure policy fidelity, intervention quality, evidence completeness, and control-placement correctness end to end.


\begin{thebibliography}{99}

\bibitem{iso42001}
ISO, ``ISO/IEC 42001:2023 --- Information technology --- Artificial intelligence --- Management system,'' Dec. 2023. [Online]. Available: \url{https://www.iso.org/standard/42001}

\bibitem{iso23894}
ISO, ``ISO/IEC 23894:2023 --- Information technology --- Artificial intelligence --- Guidance on risk management,'' 2023. [Online]. Available: \url{https://www.iso.org/standard/77304.html}

\bibitem{iso42005}
ISO, ``ISO/IEC 42005:2025 --- Information technology --- Artificial intelligence (AI) --- AI system impact assessment,'' May 2025. [Online]. Available: \url{https://www.iso.org/standard/42005}

\bibitem{iso5338}
ISO, ``ISO/IEC 5338:2023 --- Information technology --- Artificial intelligence --- AI system life cycle processes,'' 2023. [Online]. Available: \url{https://www.iso.org/standard/81118.html}

\bibitem{iso38507}
ISO, ``ISO/IEC 38507:2022 --- Information technology --- Governance of IT --- Governance implications of the use of artificial intelligence by organizations,'' 2022. [Online]. Available: \url{https://www.iso.org/standard/56641.html}

\bibitem{nistairmf}
NIST, ``Artificial Intelligence Risk Management Framework (AI RMF 1.0),'' NIST AI 100-1, Jan. 2023. doi: 10.6028/NIST.AI.100-1.

\bibitem{nistgaiprofile}
NIST, ``Artificial Intelligence Risk Management Framework: Generative Artificial Intelligence Profile,'' NIST AI 600-1, Jul. 2024. doi: 10.6028/NIST.AI.600-1.

\bibitem{nistagentinitiative}
NIST, ``AI Agent Standards Initiative,'' Center for AI Standards and Innovation (CAISI), Feb. 2026. [Online]. Available: \url{https://www.nist.gov/caisi/ai-agent-standards-initiative}

\bibitem{nistidentity}
H. Booth, W. Fisher, R. Galluzzo, and J. Roberts, ``Accelerating the Adoption of Software and Artificial Intelligence Agent Identity and Authorization,'' Initial Public Draft, National Cybersecurity Center of Excellence, NIST, Feb. 2026. [Online]. Available: \url{https://csrc.nist.gov/pubs/other/2026/02/05/accelerating-the-adoption-of-software-and-ai-agent/ipd}

\bibitem{nasa7150}
NASA, ``NPR 7150.2D --- NASA Software Engineering Requirements,'' Office of the Chief Engineer, Mar. 2022. [Online]. Available: \url{https://nodis3.gsfc.nasa.gov/displayDir.cfm?t=NPR&c=7150&s=2D}

\bibitem{nasa87398}
NASA, ``NASA-STD-8739.8B --- Software Assurance and Software Safety Standard,'' Office of Safety and Mission Assurance, Sep. 2022. [Online]. Available: \url{https://standards.nasa.gov/standard/nasa/nasa-std-87398}

\bibitem{nasahdbk2203}
NASA, ``NASA-HDBK-2203 --- NASA Software Engineering and Assurance Handbook,'' Office of the Chief Engineer, current public handbook and standards entry. [Online]. Available: \url{https://swehb.nasa.gov/}; \url{https://standards.nasa.gov/standard/nasa/nasa-hdbk-2203}

\bibitem{nist800160v2}
NIST, ``SP 800-160 Vol. 2 Rev. 1 --- Developing Cyber-Resilient Systems: A Systems Security Engineering Approach,'' Dec. 2021. doi: 10.6028/NIST.SP.800-160v2r1.

\bibitem{darpaassured}
DARPA, ``Assured Autonomy,'' Information Innovation Office program summary. [Online]. Available: \url{https://www.darpa.mil/research/programs/assured-autonomy}

\bibitem{mi9}
C. L. Wang, T. Singhal, A. Kelkar, and J. Tuo, ``MI9: An Integrated Runtime Governance Framework for Agentic AI,'' \emph{arXiv preprint arXiv:2508.03858}, Nov. 2025.

\bibitem{policyasprompt}
G. Kholkar and R. Ahuja, ``Policy-as-Prompt: Turning AI Governance Rules into Guardrails for AI Agents,'' in \emph{Proc.\ 3rd Regulatable ML Workshop, NeurIPS 2025}, \emph{arXiv preprint arXiv:2509.23994}, Nov. 2025.

\bibitem{policycards}
J. Mavracic, ``Policy Cards: Machine-Readable Runtime Governance for Autonomous AI Agents,'' \emph{arXiv preprint arXiv:2510.24383}, Oct. 2025.

\bibitem{pathpolicies}
M. Kaptein, V.-J. Khan, and A. Podstavnychy, ``Runtime Governance for AI Agents: Policies on Paths,'' \emph{arXiv preprint arXiv:2603.16586}, Mar. 2026.

\bibitem{guardrailsurvey}
Y. Dong, R. Mu, Y. Zhang, S. Sun, T. Zhang, C. Wu, G. Jin, Y. Qi, J. Hu, J. Meng, S. Bensalem, and X. Huang, ``Safeguarding Large Language Models: A Survey,'' \emph{arXiv preprint arXiv:2406.02622}, Jun. 2024.

\bibitem{foundationalguardrail}
Y. Huang, H. Hua, Y. Zhou, P. Jing, M. Nagireddy, I. Padhi, G. Dolcetti, Z. Xu, S. Chaudhury, A. Rawat, L. Nedoshivina, P.-Y. Chen, P. Sattigeri, and X. Zhang, ``Building a Foundational Guardrail for General Agentic Systems via Synthetic Data,'' \emph{arXiv preprint arXiv:2510.09781}, Oct. 2025.

\bibitem{agentsafetybench}
Z. Zhang, S. Cui, Y. Lu, J. Zhou, J. Yang, H. Wang, and M. Huang, ``Agent-SafetyBench: Evaluating the Safety of LLM Agents,'' \emph{arXiv preprint arXiv:2412.14470}, May 2025.

\bibitem{webguard}
B. Zheng, Z. Liao, S. Salisbury, Z. Liu, M. Lin, Q. Zheng, Z. Wang, X. Deng, D. Song, H. Sun, and Y. Su, ``WebGuard: Building a Generalizable Guardrail for Web Agents,'' \emph{arXiv preprint arXiv:2507.14293}, Jul. 2025.

\bibitem{gap}
A. Cartagena and A. Teixeira, ``Mind the GAP: Text Safety Does Not Transfer to Tool-Call Safety in LLM Agents,'' \emph{arXiv preprint arXiv:2602.16943}, Feb. 2026.

\bibitem{toolsafe}
Y. Mou, Z. Xue, L. Li, P. Liu, S. Zhang, W. Ye, and J. Shao, ``ToolSafe: Enhancing Tool Invocation Safety of LLM-based Agents via Proactive Step-level Guardrail and Feedback,'' \emph{arXiv preprint arXiv:2601.10156}, Jan. 2026.

\bibitem{agentdog}
D. Liu \emph{et al.}, ``AgentDoG: A Diagnostic Guardrail Framework for AI Agent Safety and Security,'' \emph{arXiv preprint arXiv:2601.18491}, Jan. 2026.

\bibitem{proofguardrail}
X. Jin, M. Duan, Q. Lin, A. Chan, Z. Chen, J. Du, and X. Ren, ``Proof-of-Guardrail in AI Agents and What (Not) to Trust from It,'' \emph{arXiv preprint arXiv:2603.05786}, Mar. 2026.

\bibitem{abcbench}
Y. Zhu \emph{et al.}, ``Establishing Best Practices for Building Rigorous Agentic Benchmarks,'' \emph{arXiv preprint arXiv:2507.02825}, Aug. 2025.

\bibitem{schneider}
F. B. Schneider, ``Enforceable Security Policies,'' \emph{ACM Transactions on Information and System Security}, vol. 3, no. 1, pp. 30--50, Feb. 2000. doi: 10.1145/353323.353382.

\bibitem{saltzer}
J. H. Saltzer and M. D. Schroeder, ``The Protection of Information in Computer Systems,'' \emph{Proceedings of the IEEE}, vol. 63, no. 9, pp. 1278--1308, Sep. 1975. doi: 10.1109/PROC.1975.9939.

\end{thebibliography}
\end{document}